\documentclass{article}
\usepackage{arxiv}

\usepackage{times}
\usepackage{color,soul}
\usepackage{url}
\usepackage{hyperref}
\usepackage[utf8]{inputenc}
\usepackage[small]{caption}
\usepackage{caption}
\usepackage{graphicx}
\usepackage{amsmath}
\usepackage{bm}
\usepackage{amsthm}
\usepackage{booktabs}
\usepackage{algorithm}
\usepackage{algorithmic}
\usepackage{enumitem}
\usepackage[font=footnotesize]{subcaption}
\usepackage{natbib}
\usepackage{flushend}

\newcommand{\new}[1]{{\color{black}#1}}

\title{Sensor Control for Information Gain in Dynamic, Sparse and Partially Observed Environments}

\author{ 
    {J. Brian Burns}\thanks{All authors have contributed equally}\\
	Artificial Intelligence Center\\
	SRI International\\
	Menlo Park, CA, 94025 \\
	\texttt{brian.burns@sri.com} \\
	\And
	{Aravind Sundaresan} \\
	Artificial Intelligence Center\\
	SRI International\\
	Menlo Park, CA, 94025 \\
	\texttt{aravind.sundaresan@sri.com} \\
	\And
	{Pedro Sequeira} \\
	Artificial Intelligence Center\\
	SRI International\\
	Menlo Park, CA, 94025 \\
	\texttt{sequeira@ai.sri.com} \\
	\And
	{Vidyasagar Sadhu} \\
	Artificial Intelligence Center\\
	SRI International\\
	Menlo Park, CA, 94025 \\
	\texttt{srikanthvidyasagar.sadhu@sri.com} \\
}

\date{}

\renewcommand{\headeright}{}
\renewcommand{\undertitle}{A preprint}

\hypersetup{
pdftitle={Sensor Control for Information Gain in Dynamic, Sparse and Partially Observed Environments},
pdfsubject={
We present an approach for autonomous sensor control for information gathering under partially observable, dynamic and sparsely sampled environments. We consider the problem of controlling a sensor that makes partial observations in some space of interest such that it maximizes information about entities present in that space. We describe our approach for the task of Radio-Frequency (RF) spectrum monitoring, where the goal is to search for and track unknown, dynamic signals in the environment. To this end, we extend the Deep Anticipatory Network (DAN) Reinforcement Learning (RL) framework by (1) improving exploration in sparse, non-stationary environments using a novel information gain reward, and (2) scaling up the control space and enabling the monitoring of complex, dynamic activity patterns using hybrid convolutional-recurrent processing. We also extend this problem to situations in which taking samples from the intended RF spectrum/field is expensive and therefore limited, and propose a model-based version of the original RL algorithm that fine-tunes the controller using a model of the environment that is iteratively improved from limited samples taken from a simulated RF field. Results in simulated RF environments of differing complexity show that our system outperforms the standard DAN architecture and is more flexible and robust than baseline expert-designed agents. We also show that our approach is adaptable to non-stationary emission environments.
},
pdfauthor={J. Brian Burns, Aravind Sundaresan, Pedro Sequeira, Vidyasagar Sadhu},
pdfkeywords={Sensor Control, Information Gain, Reinforcement Learning, Partial Observability, Dynamic Environment, Sparse Rewards},
}

\begin{document}

\maketitle

\begin{abstract}
We present an approach for autonomous sensor control for information gathering under partially observable, dynamic and sparsely sampled environments that maximizes information about entities present in that space. We describe our approach for the task of Radio-Frequency (RF) spectrum monitoring, where the goal is to search for and track unknown, dynamic signals in the environment. To this end, we extend the Deep Anticipatory Network (DAN) Reinforcement Learning (RL) framework by (1) improving exploration in sparse, non-stationary environments using a novel information gain reward, and (2) scaling up the control space and enabling the monitoring of complex, dynamic activity patterns using hybrid convolutional-recurrent neural layers. We also extend this problem to situations in which sampling from the intended RF spectrum/field is limited and propose a model-based version of the original RL algorithm that fine-tunes the controller via a model that is iteratively improved from the limited field sampling. Results in simulated RF environments of differing complexity show that our system outperforms the standard DAN architecture and is more flexible and robust than baseline expert-designed agents. We also show that it is adaptable to non-stationary emission environments.
\end{abstract}



\section{Introduction}\label{sec:intro}

\textbf{Overview.} 
Sensor control that maximizes information gain under partially observable and dynamic environments is an important problem that has several applications~\citep{Satsangi2020}. For example, consider the problems of tracking widespread activity from a limited, but controllable field of view, or tracking carbon monoxide levels over a geographic area with the minimum number of sample sites.
A relevant problem, which we study in this work, is Radio Frequency~(RF) spectrum monitoring, which involves detecting and tracking multiple dynamic signals in a potentially large RF spectrum using an RF receiver (sensor) with only a limited, but tunable, reception band. 
All these problems require a sequential decision making approach that selects the sensor(s) or sensor settings for each time instant, based on past observation history, to maximize information gain e.g., knowledge about signal activity.

\noindent\textbf{Challenges.} The above tasks present several challenges: (i)~\textit{partial observability}: each sensor (or sensor band) can only provide partial observations of the underlying state; (ii)~\textit{dynamic environments}: the underlying environment is stochastic and could be non-stationary, making it hard to track its state over time; (iii)~\textit{sparse environments}: it is possible that useful information may only be infrequently collected, meaning that the majority of observations are mostly non-informative; \new{(iv)~\textit{costly samples}: taking samples from an actual fielded sensor is costly and so interactions with it are restricted according to some \emph{budget}.}

\noindent\textbf{Approach.} 
In this paper, we study and evaluate our approach for an abstracted RF spectrum monitoring task. In particular, the goal is to control a band-limited RF receiver to optimize multi-signal detection and tracking throughout an extensive RF environment. The autonomous controller is given a range of the spectrum (discretized into several frequency bands) within which to operate. However, it is not given information about the specific frequencies, densities or distributions of the signals within this range. 
For purposes of experimentation, we developed an RF simulator on which a controller is trained to decide 
\emph{which} band to sample at each instant to accomplish the goal of tracking/predicting the signals in the spectrum. During the training process, the controller learns the general behavior of the signals,  such as frequency switching patterns and typical durations, and learns how to use this knowledge to search for and track these types of signals in novel situations.

Towards this goal, we propose an approach based on the Deep Anticipatory Network (DAN)~\citep{Satsangi2020} framework for optimal sensor control in information gathering tasks. DAN uses deep Reinforcement Learning (RL) with prediction rewards, where the action suggested by a \emph{value-network}, that implements the value function, is rewarded whenever a separate \emph{model-network} makes a correct prediction of the true state based on observations that are the result of the action. In order to scale to large problem spaces, e.g., large spectrum and/or small reception bandwidth, and to efficiently learn and process signals with potentially complex frequency-time patterns, we extended the DAN approach by implementing the prediction and control functions using neural networks with hybrid convolutional-recurrent layers, such as the ConvLSTM layer~\citep{xingjian2015convolutional}. Additionally, as the environment is dynamic and sparse, we propose novel information-gain rewards for our RL approach to encourage exploration thereby avoiding sampling regions of spectrum where signals were already identified.

Another important aspect of our approach, that was not considered in the original DAN work, is the use of potentially limited data from a fielded controller (referred to as \emph{experience feedback}) to update the RF simulator used for training.
In particular, a controller is trained in simulated RF environments by using a simulator that parameterizes a distribution over several aspects controlling the dynamics of the RF spectrum, e.g., the number of emitting signals, the frequency at which they emit, different communication patterns, etc. We refer to this as the \emph{lab} simulator. Using it, the controller can train on many and diverse samples of environments --- far more than can be sampled out in the actual field environments over a reasonable training period. However, in a real-world situation, the distribution over the actual field RF environments of interest may be non-stationary, which can make the
conditions in which the controller was trained obsolete and consequently the controller sub-optimal. In this work, we investigate how the lab simulator can be updated using field experience data gathered by deploying the trained controller in the field for a limited amount --- the interactions are restricted according to some budget. These field samples are used to adjust the lab simulator's parameters such that the generated environments closely match those encountered in the field, and the controller's policy can be fine-tuned by RL-based retraining on the updated lab simulator. We study this problem by using an additional \emph{field} simulator --- one that holds the ground-truth of the distribution over the environment parameters --- to simulate the collection of limited field data and to test the retrained system.

In this work, we will refer to a stochastic model of an RF environment, which determines its dynamics, as a \emph{spec}. The field simulator will have a \emph{field spec}, which is hidden to the system, and during any given training 
iteration, the lab simulator will have a \emph{lab spec} that is used to train the controller.
\new{In addition to estimating the field parameters/spec using field samples, we study an alternative approach of learning the field state dynamics model directly using field samples. We then use this field state model in the lab as an environment to train a controller using model-based RL. 
}

\noindent\textbf{Contributions.} Our contributions in this paper are as follows:
\begin{itemize}[noitemsep,nolistsep,leftmargin=*]
    \item Extending the DAN framework to handle the challenging problem of monitoring an RF spectrum.
    \item Scaling up the control space and enabling the monitoring of complex, dynamic activity patterns using hybrid convolutional-recurrent processing steps.
    \item Improving exploration in sparse, non-stationary environments using novel information gain rewards.
    \item A supervised learning variant that is shown to perform better than RL approaches in certain environments.
    \item Two model-based RL approaches capable of model updating using experience feedback from limited field deployment.
    \item An evaluation of the the proposed approaches using various architectural configurations showing their benefit over baseline controllers in different environments. 
\end{itemize}


\section{Related Work}\label{sec:rel-work}


\noindent\textbf{General approaches.} Information gathering under partial observability, which is related to Partially Observable Markov Decision Processes (POMDPs) \citep{Kaelbling1998}, has been explored in several domains. For example, \citet{Satsangi2020}, present a technique (our baseline approach) for information gathering using Deep Anticipatory Network~(DAN) that decides which camera to switch on to track people in a mall. They also applied a related approach to active perception in robotics~\citep{satsangi2018exploiting}. \citet{Wang2018} propose an RL technique for information gathering using sparse mobile crowd-sensing that tells which cell to collect data from to minimize the uncertainty in state estimation. \citet{Mnih2014} and \citet{haque2016recurrent} present an image classifier that adaptively selects regions for processing at high resolution. 
Most of these techniques do not address our problem combining dynamic environment, reward sparsity and limited  field experience. In the domain of generic POMDP solutions, \citet{james2009sarsalandmark} present an algorithm for learning in POMDPs assuming availability of `landmarks' or special states that are not hidden to the agent, but not available in our domain. 
\citet{katt2017learning} present an approach based on Monte-Carlo tree search and look-ahead planning for solving POMDP problems in an online manner. However, due to large dimensional state space, such an approach is not scalable to our domain.

\noindent\textbf{RF domain.} In the domain of RF spectrum monitoring, most of the existing work is on signal detection and identification~(SDI) rather than tracking. \citet{Franco2020} present a hierarchical two-step approach for SDI.
At a finer scale, they first use a sweeping window to detect the local presence of signals, and at a larger scale, these local detections are integrated using a frequency-time region proposal network.
\citet{Kulin2018} present a single-step, an end-to-end deep learning approach for wireless signal detection.
\citet{Mendis2019} present an attention-driven RL technique for signal detection in a wide-band spectrum. The proposed method consists of two main components: a spectral correlation function (SCF) based spectral visualization scheme and a spectral attention-driven RL mechanism that adaptively selects the spectrum range and implements the intelligent signal detection. This approach however assumes that the modulation technique is known \emph{a priori}.

\begin{figure}[!t]
    \begin{center}
    \includegraphics[width=0.5\linewidth]{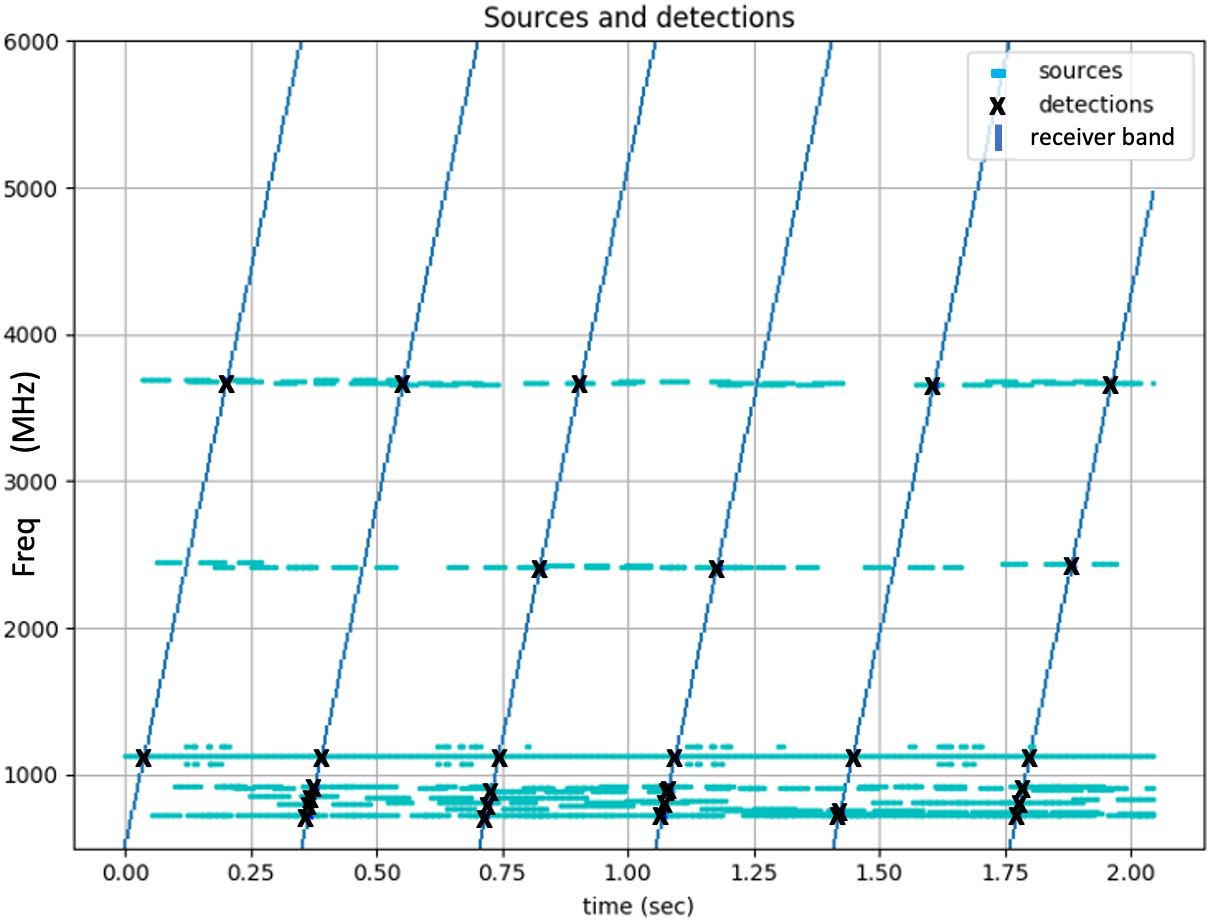}
    \end{center}
    \caption{A partially observed RF environment with sequential scan for control, which misses the short signal bursts at $\mathrm{1.2}$GHz. Plot: time(secs) vs frequency(MHz); cyan: signals; dark blue: sampled bands; black x marks: detected signals.}
    \label{fig:rf_env_example}
\end{figure}

\noindent\textbf{Experience Feedback.}
This problem is related to other types of machine learning tasks. In transfer learning for RL, (\citet{Taylor2009} and \citet{Zhu2020}) there is a source (lab spec) and target (field spec) population; however, partial observability is not addressed. In model-based value expansion (imagination rollouts), the number of real (field) interactions is reduced by doing additional training with updated simulator (\citet{Feinberg2018}). \citet{Kalweit2017} and \citet{Hafner2020} discuss RL imagination rollout approaches based on uncertainty and latent space. However, unlike our problem, most of these works deal with continuous actions. Model-based RL traditionally involves interactions between a planning module and RL training (\citet{Moerland2020}); \citet{Li2020} discuss a technique for accelerating model-free RL using imperfect models for the related application of spectrum access. \citet{Hua2019} present a GAN-based method to learn the state-action values for resource management in network slicing. \citet{Huang2021} present a Generative Adversarial Interactive Reinforcement Learning that combines the advantages of GAN-based learning and interactive RL. In our problem, the field state to be learned is a time-series system and most existing approaches do not handle this case. 

\section{Proposed Solution}\label{sec:prop-soln}

\subsection{State Estimation and Tracking }\label{sec:prop-soln:dan}
The problem involves finding and tracking multiple signals that have activity patterns of varying complexity involving several unknown parameters. 
A controller agent is trained and tested using multiple episodes (trials), where each episode has a new signal environment sampled from the same stochastic environment model (\emph{spec}), and the episode involves a sequence of control (frequency selection) and observation (signal detection) steps. The specific signal patterns of each new environment are 
unknown to the agent and must be inferred through tracking and monitoring during the interaction.
To solve this problem, we extend the original DAN architecture and rewards.

\noindent\textbf{Problem Formulation.} Following the DAN approach in \citep{Satsangi2020},
we consider a Partially Observable Markov Decision Process (POMDP) \citep{Kaelbling1998} to model the dynamics of our RF environment.
We denote $s \in S$ to represent the hidden state of the environment, $y \in Y$ to denote a target variable of interest that depends only on $s$, $z \in \Omega$ to denote a partial observation that is correlated with $y$ and $a \in A$ to denote the action taken by the agent. At each discrete timestep, $t$, the agent takes an action $a^t$, the environment transitions to state $s^{t+1}$ and the agent receives an observation $z^{t+1}$. The goal of the agent is to correctly predict the target variable $y^t$ given the history of previous actions and observations denoted by $h^t = (a^0, z^1, ..., a^{t-1}, z^t)$. We denote by $\hat{y}$ the agent's prediction of $y$. At each step, the agent receives a reward, denoted by $r(y^t, \hat{y}^t)$, that indicates how similar $y^t$ and $\hat{y}^t$ are.

\noindent\textbf{RF Environment.} In the context of RF spectrum monitoring, $s$ represents the state of the whole RF spectrum, 
that contains a set of unknown \emph{entities} interacting with each other and transmitting signals spread across frequency and varying in time. Further, $y$ is the vector representing signal activity ($0$ for none and $1$ for activity) at all frequency bands, $a$ represents the frequency band(s) that the 
agent samples, and $z$ represents observed detections (the presence/absence of signals at the sampled bands); $h$ is the history of sampled frequencies and observations. The agent's actions can only sub-sample the spectrum at specific frequencies. The goal of the agent is to select actions at each timestep in order to accurately report the presence/absence of signals in \emph{all} frequency bands at that timestep.
Fig.~\ref{fig:rf_env_example} shows an example of a partially observed RF environment. One or more bands can be active (contain signals) at any given time. Here, a sequential scan is performed across the spectrum, which misses the short signal bursts around $1{,}200$MHz.



\noindent\textbf{Environment Spec Details.} An environment spec defines an environment population model as a range of possible values for 
the following parameters: (i)~\emph{number}: number of interacting signal pairs; (ii)~\emph{width}: total number of spectral bands spanned by the pairs; (iii)~\emph{period}: period of the signal pair interaction; (iv)~\emph{duty cycle}: duration of one signal over the other during an interaction cycle; 
(v)~\emph{frequency}: the frequency of the lowest of the signal pair;
(vi)~\emph{start}: timestep when signal pair appears in spectrum. Table~\ref{tab:envs} lists some of the environment specs used in our experiments.

\begin{figure}[t]
    \begin{center}
    \includegraphics[width=3.5in]{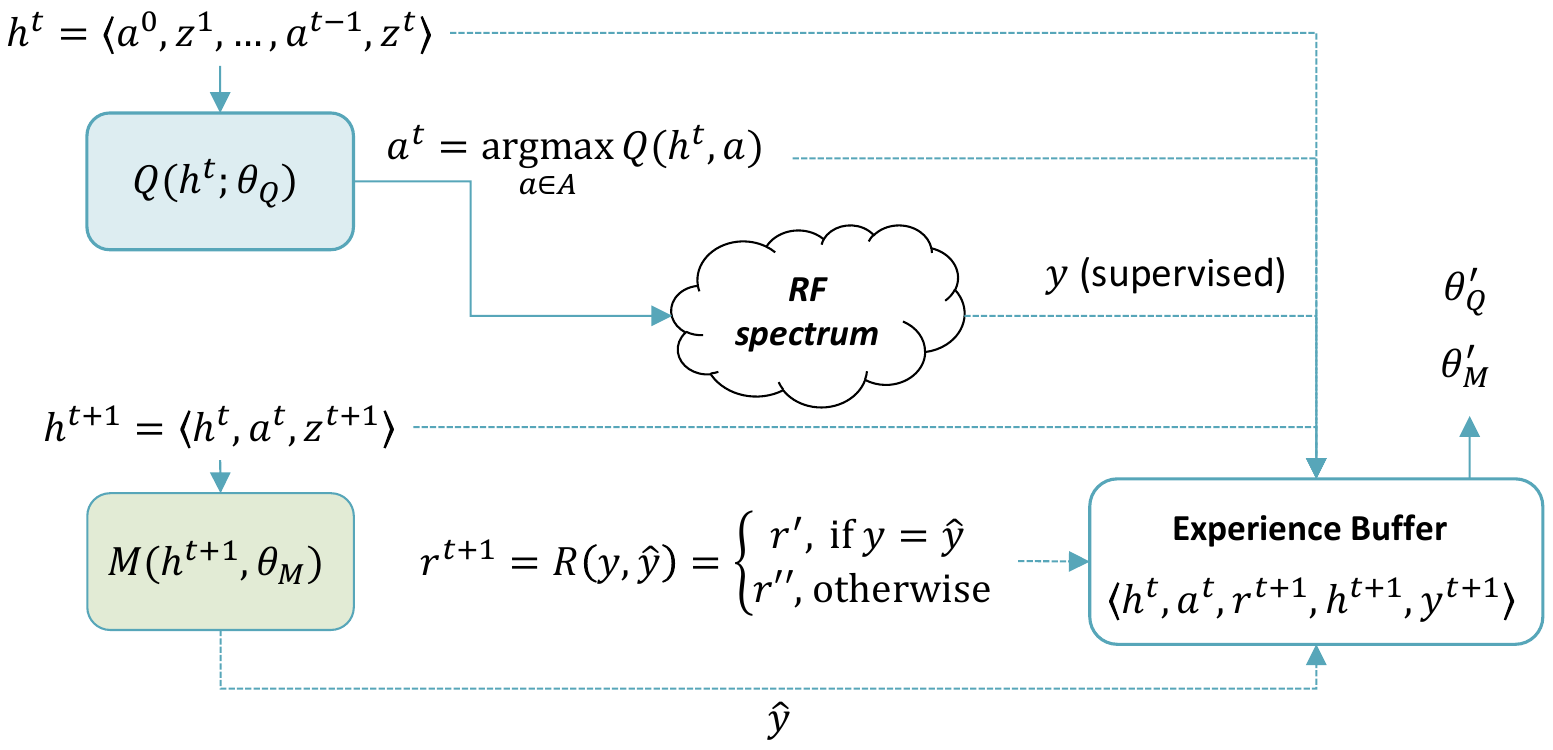}
    \end{center}
    \caption{Deep Anticipatory Network~(DAN) overview.}
    \label{fig:dan}
\end{figure}

\begin{figure}[t]
    \begin{center}
    \includegraphics[width=0.7\linewidth]{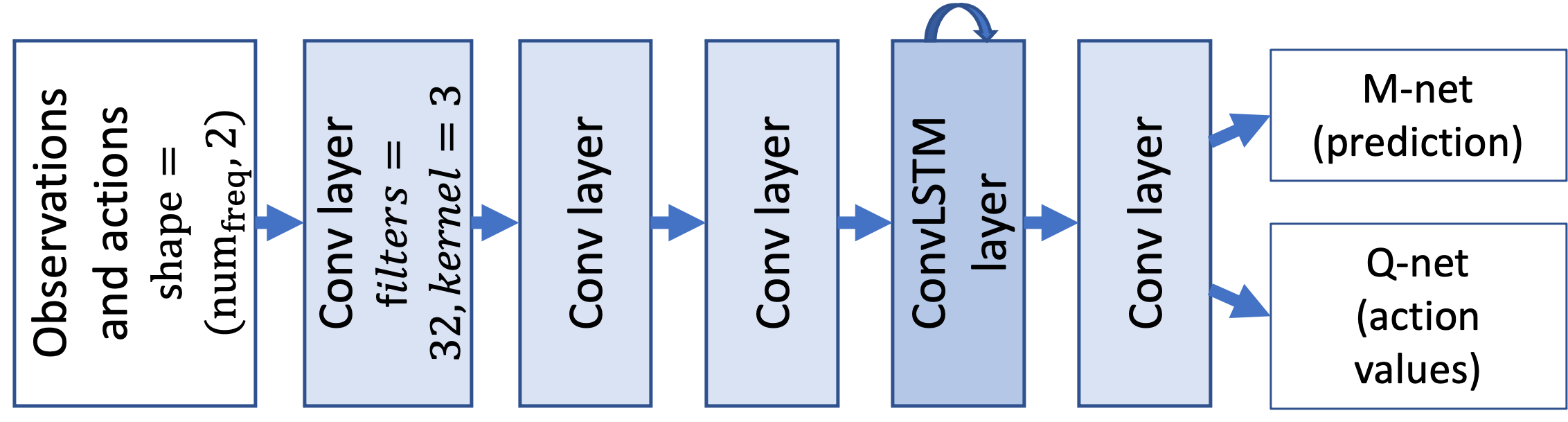}
    \end{center}
    \caption{ConvLSTM-DAN architecture with Q and M outputs. The ConvLSTM is convolutional in frequency and recurrent in time.}
    \label{fig:convlstm_dan}
\end{figure}

\subsection{DAN framework} 
Deep Anticipatory Networks~(DANs)~\citep{Satsangi2020} were developed for infor\-mation-gathering tasks under partial observability e.g., tracking people using cameras. 
A DAN makes control decisions that maximize the information gathered (minimize the uncertainty) about a target variable of interest $y$ that cannot be fully-observed. The key point is that, for this type of task, DAN avoids complex belief state updating and instead uses state prediction rewards to guide the agent behavior.%
\footnote{This has been demonstrated in~\citet{Satsangi2020} to be equivalent in effect to belief updating.} 
DAN uses two neural networks to train a policy, as shown in Fig.~\ref{fig:dan}. The first, $\boldsymbol{Q}$, takes the history of previous actions and observations, $h_t$, and produces $Q$-values for the different actions. 
A second network, $\boldsymbol{M}$, takes the history $h_t$ plus the last action and observation to predict $\hat{y}$, i.e., an estimate of target variable $y$. $M$ is trained in a supervised manner with ground truth data, i.e., using the actual target values $y$. $Q$ is a standard value function network trained using RL but here it is rewarded when $M$
's predictions are accurate, i.e., when $\hat{y}$ is similar to $y$. $Q$ is trained to maximize cumulative discounted reward, $\sum_t{r_t\gamma^t}$. The $Q$ and $M$ networks are trained simultaneously and, after training, only $Q$ is used for autonomous control for information gain. In all of our DAN variants, $M$ is trained using weighted binary cross entropy
(WBCE) as a loss, weighted to emphasize our metrics; see the discussion on rewards and metrics in Sec.~\ref{sec:prop-soln:dan:rewards}.

\subsubsection{ConvLSTM, Predictive and InfoMax DAN Architectures} 
The problem of RF spectrum monitoring presents several challenges, including: being able to scale to large discrete frequency and time domains, identifying repeating signal patterns in frequency-time space and coping with non-stationary nature of the environment. The original 
DAN does not address these challenges; we propose various architectural and reward modifications to address them.

\noindent\textbf{ConvLSTM-DAN.} 
The baseline DAN uses fully connected/dense layers for both $Q$ and $M$ networks with Rectified Linear Unit (ReLU) activation, which doesn't scale well and makes learning of frequency-invariant activity patterns difficult. We address the scalability of the problem space by using convolutional layers (convolution in frequency). We also introduce hybrid convolutional-LSTM (Long Short Term Memory) / ConvLSTM layers~\citep{xingjian2015convolutional}, which combine convolution in frequency dimension and recurrence/memory in time dimension for learning translation-invariant frequency-time activity patterns. This architecture is shown in Fig.~\ref{fig:convlstm_dan}. For ConvLSTM training, we use a reward that is a function of the intersection over union (IoU) of the predicted output from $M$ and the ground truth; see the discussion on rewards and metrics in Sec.~\ref{sec:prop-soln:dan:rewards}.
As the input to both $Q$ and $M$ networks are similar, and action selection and state prediction are related tasks, the layers used to compute features can be shared to reduce the training space. We call this enhanced framework \emph{ConvLSTM-DAN}, which has two outputs, one for $Q$ and the other for $M$. (All of our DAN variants discussed in this paper use a shared architecture.)
After studying the effect of different well-known DQN/RL enhancements, we found that Dueling DQN~\citep{Wang2016} improved results across the board. Therefore, our standard ConvLSTM-DAN framework incorporates it (not shown in Fig.~\ref{fig:convlstm_dan}): for $Q$, the output of the last convolutional layer is split into two streams---Value and Advantage---which are then combined to give the $Q$ output. Double DQN~\citep{Vanhasselt2015} was not considered as useful and was not incorporated.


\noindent\textbf{Predictive DAN.} 
If the environment spec defines a variable number of signals 
it introduces an interesting exploration / exploitation challenge with subtle differences in the payoff odds between carefully tracking the signal found vs. finding a new one.
This problem is 
exacerbated in reward-sparse scenarios (with an empty spectrum most of the time). In order to encourage exploring an unknown state space and at the same time exploit the known state space, we propose an enhancement over the ConvLSTM-DAN that provides an auxiliary predictive reward, which we refer to as the Predictive-DAN. 
This is similar to the Intrinsic Curiosity Module~\citep{pathak2017curiosity}. In Predictive-DAN, the shared network in Fig.~\ref{fig:convlstm_dan} has 3 outputs: $m$-output, $M=y_t | h_t$ (the current state given current observation history), $q$-output, $Q=a_{t+1} | h_t$ (next action to take given current observation history) and a predictive output, $\bm{P}=y_{t+1} | h_t$ (the next state given current observation history). Here, state refers to target variable, $y$. When the next action, $a_{t+1}$ is actually taken, then the m-network estimates the current state given current information: $y_{t+1} | h_{t+1}$, where $h_{t+1} = \{h_t, a_{t+1}, z_{t+1}\}$. We can denote the information gained from this action as $infogain = mean(abs(y_{t+1} | h_t - y_{t+1} | h_{t+1}))$, which is used as an auxiliary reward to train the Predictive-DAN $Q$-network: $IoUreward + infogain * 10$. In practice, typically only the observed `band' will change if at all, but if the network has learned relations between signals in other bands, that could be affected too. This is similar to the reward that can be used when there is no ground-truth (the difference being that this is only computed for the observed state and not full state).

\noindent\textbf{InfoMax-DAN.} While in above architectures, we compute and use information gain only for the executed action, it is also possible to compute the information gained for all actions instead of just the action taken. We use this idea to train $Q$-net in a fully supervised manner without using RL. This approach maximizes information gained in a single step (with no cumulative reward). We refer to this alternative approach as InfoMax-DAN, which is fully supervised version of Predictive-DAN.

\subsubsection{Rewards and Metrics}\label{sec:prop-soln:dan:rewards}

Since the goal in our control problem is to maximize information about signal presence, it seems reasonable to start with the commonly used detection/localization metric Intersection over Union~(IoU). In our case, \emph{intersection} is the count of (time, frequency) positions where true signals coincide with predicted ones, and \emph{union} is the total count of true positions plus the total for predicted ones (where predictions can either be probabilities or thresholded, binary states). We use the following variants of IoU for rewards and metrics in our experiments: \textit{Instantaneous IoU} is IoU for one slice of time; \textit{Cumulative IoU} is cumulative IoU up to a given time; \textit{Block IoU} is cumulative for the last $N$ timesteps. The Instantaneous IoU can provide frequent rewards but is unstable in sparse environments, which be can addressed using a differential reward: \textit{Differential Block IoU} $ = BIoU_t^{N+1} - BIoU_t^N$, where the $BIoU$ are block IoU, for blocks of fixed $N$.
For losses to train prediction, we use weighted binary cross entropy (WBCE), which is an effort to tilt the commonly used BCE more towards the IoU-based reward functions: the error for existing signals is weighted more than the error for non-existing signals. (The weight for positions \emph{without} signals is set to the expected signal density, $0.1$. ) In experiments, we have noticed that WBCE is qualitatively better than BCE or IoU.

\subsection{Experience-feedback via Model-based RL}\label{sec:prop-soln:exp-fb}

Often, the \emph{lab spec} used for training does not entirely cover the field (ground-truth) population of environments --- for example, the domain knowledge employed may be outdated. To address this challenge, we introduce an experience feedback loop. Once trained on the lab spec, the agent is then deployed in field environments (here, simulated by a \emph{field spec}) for a limited time, during which it collects samples that might be sparse. In this fielded phase, unlike the lab training, the agent does not have access to the underlying full state required to train the $M$ network via prediction loss. Rather, these experiences are used to estimate the environments' dynamics, which are then used to update a lab spec/model. The agent is then retrained using the updated spec/model and re-deployed for more samples, 
and the process repeats for a number of times. 
We develop two model-based RL approaches to implement this feedback loop: (i)~\textit{field spec estimation} uses collected experiences to estimate the field spec parameters
directly
, which is then sampled to retrain the controller; (ii)~\textit{field state estimation} uses them to build a deep, generative model of the field state sequence population, from which we sample to retrain the controller.

\subsubsection{Field Spec Estimation}


Experience feedback using field spec estimation uses the following basic loop: (i)~deploy a controller in the field (simulator with field spec), (ii)~use an expert system to estimate the field parameters from the resulting samples (Sec.~\ref{sec:perf-eval:compare}), (iii)~add the resulting \emph{estimated} field spec to a pool of training specs (including original lab spec), (iv) train the DAN controller using samples from the training pool of specs, and (v) deploy the resulting trained DAN. We studied three different variations of these steps. \textbf{Estimating Spec from Feedback:} in step (i), deploy our expert controller (Sec.~\ref{sec:perf-eval:compare}), and in step (iv), train from only the estimated field spec. \textbf{Estimating Spec for Retraining:} in step (i), deploy a DAN controller trained on a general lab spec, and in step (iv), train from the spec pool with sample rates: estimated field spec=$0.7$, lab spec=$0.3$ (to avoid forgetting). \textbf{Bootstrapping:} do multiple iterations of the latter scheme, varying the field spec each deployment and training from all previously estimated specs.

\begin{figure}
\begin{center}
\includegraphics[width=1\linewidth]{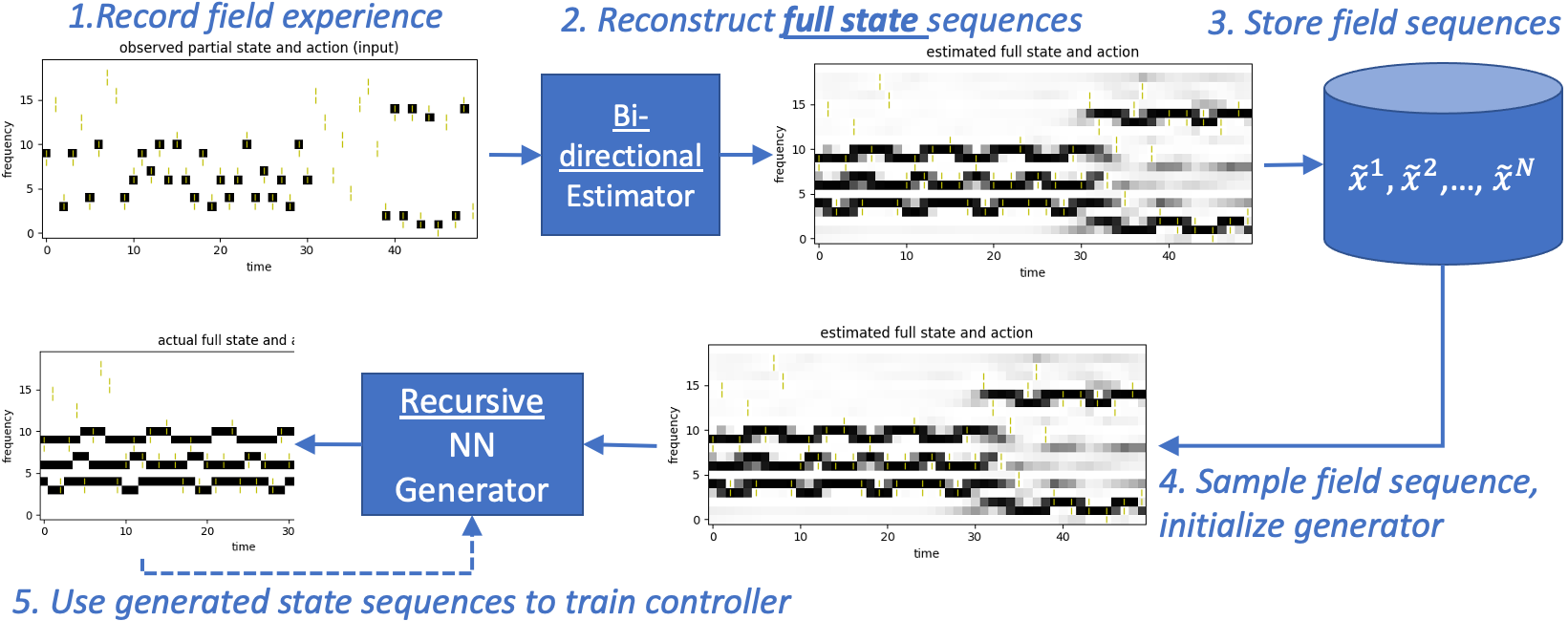}
\end{center}
\caption{\new{Experience feedback loop via direct field state estimation.}}\label{fig:model_based1}
\end{figure}

\subsubsection{Field State Estimation}\label{sec:prop-soln:exp-fb:state}

One drawback of the Field Spec Estimation approach is that it assumes that observed dynamics in the field can be modeled by fitting a predetermined set of (known) simulator parameters.
In realistic settings, we cannot anticipate all parameters governing the observed behavior in the RF spectrum --- e.g., what if the entities are using a different communication protocol? 
To address this challenge, we follow a different approach to experience feedback that retrains the agent using Machine Learning (ML)-based methods that do not rely on parameterized environment specifications. Instead, the approach discussed here trains the controller on full, extended state sequences reconstructed and extrapolated from partially observed field samples. This will be evaluated against one that directly does controller fine-tuning on the raw, partially observed samples.

In \emph{field state estimation}, we first train our DAN controller using a generic Lab Spec. We then adopt the following procedure (Fig.~\ref{fig:model_based1}): (i)~deploy and collect field experiences, which are sequences of partially observed states; (ii)~reconstruct full state sequences from the partially observed samples using a bidirectional Recurrent Neural Network~(RNN) or U-net architecture~\citep{Ronneberger2015}; (iii)~store reconstructed field sequences in a database; (iv)~sample these stored sequences and use to initialize an RNN-based generator that then emits extended sequences; (v)~binarize the generated sequences 
($1$=signal, $0$=none); (vi)~use these to train the controller, instead of a parameterized spec.

This will be compared to a DAN controller that is fine-tuned in the field directly (no lab retraining) using the available \textit{partially} observed field states (see 1. in Fig.~\ref{fig:model_based1}). As we have ground truth only in the particular band sampled, the reward obtained by the $M$-network was modified to be only the prediction in this band.

\section{Performance Evaluation}\label{sec:perf-eval}

\begin{figure}[!t]
    \centering
    \begin{subfigure}[b]{0.49\linewidth}
        \includegraphics[width=\linewidth]{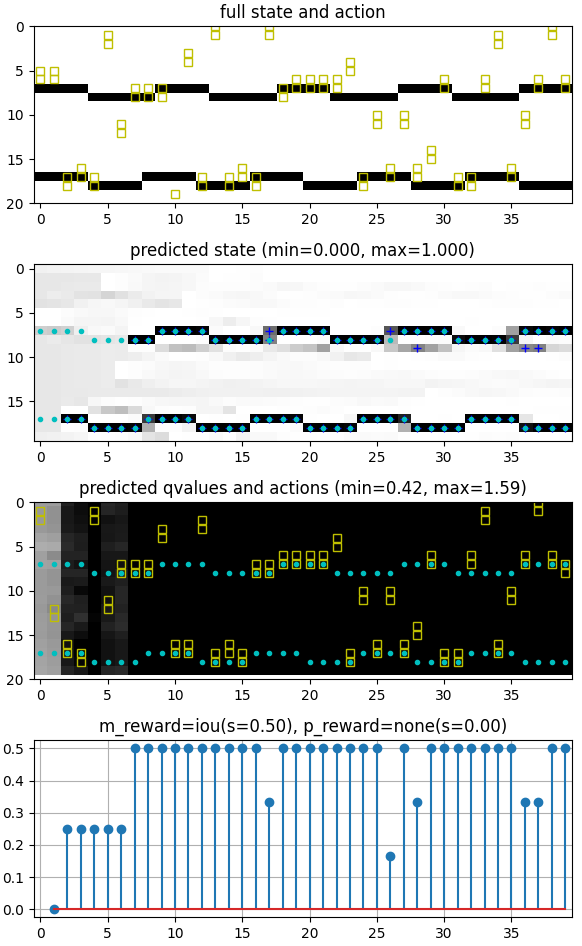}
        \caption{SpecA}%
	    \label{Fig:demo_specA}
    \end{subfigure}
    \begin{subfigure}[b]{0.49\linewidth}
        \includegraphics[width=\linewidth]{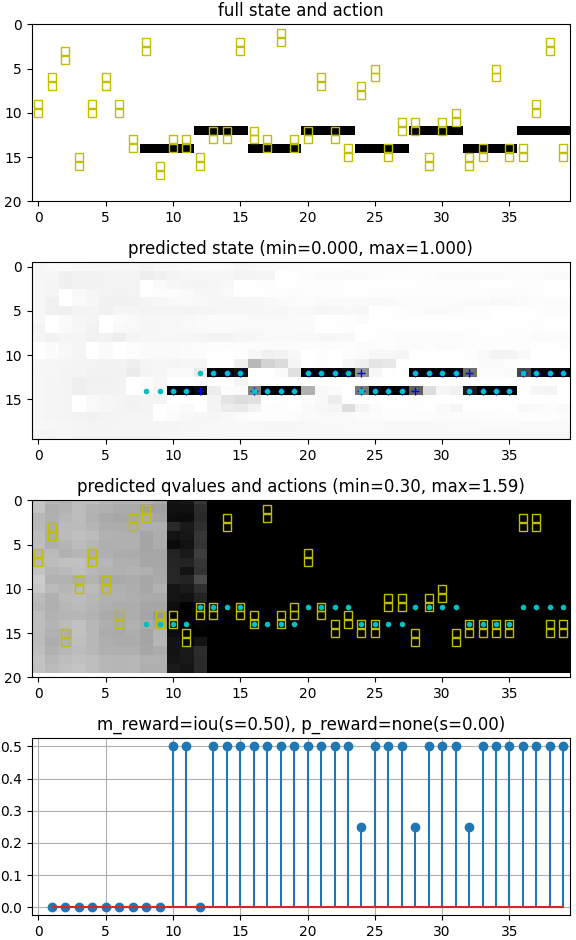}
        \caption{SpecB2}%
	    \label{Fig:demo_specB2}
    \end{subfigure}
    \caption{Training episodes from SpecA (a) and SpecB2 (b). First row: the full state (bands with signal activity are black and actions taken are gold squares). Second row: predicted state ($\mathbf{\textit{M}}$-net output) with signal probability in gray-scale (white = $\mathbf{0}$, black $\mathbf{>0.5}$) and blue dots are real signal. Third row: $\mathbf{\textit{Q}}$-net output (q-values) in gray-scale with action taken (gold boxes) and true state (blue dots). Fourth row: IoU rewards per step, where IoU measures predicted state -- true state match.) } 
    \label{fig:demo_result}
\end{figure}

\textbf{Experimental Setup.}
In order to have sufficient complexity in environment and control space to test our proposed approach, we considered environments that vary in terms of the number of communicating signal pairs and their properties; Table~\ref{tab:envs} shows some of the environment specs studied.
We will start in Section~\ref{sec:perf-eval:compare} by comparing our DAN  controller with expert-designed systems and study their adaptability using
SpecA as the environment for which the expert-designed systems are optimized, while SpecB1 and SpecB2 -- environments with increased variation and complexity -- are used to test the adaptability. Example episodes from SpecA and SpecB2 are shown in Fig.~\ref{fig:demo_result}.
Then, in Sec.~\ref{sec:perf-eval:versions}, we will extend the study to include non-stationary, semi-periodic and wider spectrum environments with  multiple signal classes, such as in Fig.~\ref{fig:rf_env_example}.


\begin{table}[tbp]
    \small
    \centering
    \begin{tabular}{@{}l r r r  r r r r r@{}}
    \toprule
    \textbf{Params} & \textbf{A} & \textbf{B1} & \textbf{B2}  & \textbf{C1} & \textbf{C2}  & \textbf{F1} & \textbf{F2} & \textbf{F3} \\ 
    \midrule
    \emph{Number}     & $2$      & $[1,2]$ & $[1,2]$  & $[1,2]$ & $2$ & $1$ & $2$ & $[1,2]$  \\
    \emph{Width}   & $2$      & $[2,3]$ & $[2,3]$  & $3$ & $2$  & $3$ & $2$ & $[2,3]$\\
    \emph{Period}     & $[8,9]$ & $[8,9]$ & $[8,9]$  & $[8,9]$ & $[6,9]$  & $[8,9]$ & $[8,9]$ & $[6,7]$\\
    \emph{Dutycyc} & $4$      & $[4,5]$ & $[4,5]$  & $4$ & $[2,5]$ & $4$ & $7$ & $[3,4]$\\
    \emph{Freq}     & Rand & Rand & Rand     & Rand & Rand & Rand & Rand & Rand\\
    \emph{Start} & $0$ & $0$ & $[0,10]$         & $0$ & $0$ & $0$ & $[5,10]$ & $[0,5]$\\
    \bottomrule
    \end{tabular}
    \caption{Some environment specs considered. The parameters are defined in Sec.\ref{sec:prop-soln:dan} (Environment Spec Details); Rand means random for each signal pair; [a,b] means random inclusively within range [a,b].
    }\label{tab:envs}
\end{table}



\subsection{Comparison with non-ML Controllers}\label{sec:perf-eval:compare}
We hand-coded four controllers that follow simple, but often effective, behavior rules for RF environments to serve as baselines to evaluate the DAN-based system robustness:
%
\begin{description}
    \item[Random:] randomly selects bands in the spectrum. For prediction, it uses a \emph{persistent state} scheme: each band's signal state (active/inactive) is set to the last observation in it. Initially, all bands are considered inactive.
    %
    \item[Scan:] sequentially selects bands in the spectrum, 
    uses the same persistent state scheme for prediction.
    \item[Scan-and-Dwell:] estimates the different spec parameters governing the dynamics of the environment from the sampled observations. The controller is given ranges for the parameters but does not know their true values. It traverses the spectrum sequentially and whenever an action triggers a signal detection, it subsequently samples  the same band until the true parameters controlling the dynamics in \emph{that} band are learned, via a process of elimination. The predicted state is computed for each from the estimated parameters. 
    \item[Expert:] uses the same process as the above controller to estimate the spec parameters from experience. At each step, it selects the band with the highest associated uncertainty, i.e., whose current range of possible values is the largest. 
\end{description}

In particular, the Scan-and-Dwell and Expert controllers serve as reasonable upper bounds on performance since in our experiments they are given relatively small ranges for the different spec parameters around their ground-truth values. In other words, an ML agent that has been trained in environments whose spec parameters are similar to those governing the environment on which it is being evaluated can be expected, at best, to attain a performance similar to that of the best hand-coded controller.

\begin{figure}[tbp]
    \centering
        \begin{subfigure}[b]{0.5\textwidth}
            \centering
            \includegraphics[width=2.8in]{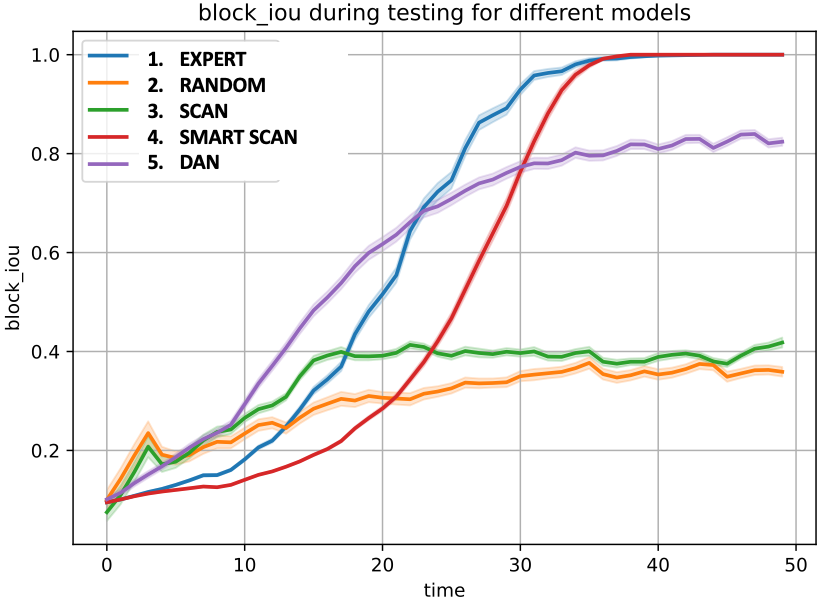}
            \caption{SpecA environments}
            \label{fig:man_dan_perf_specA}
        \end{subfigure}%
        \begin{subfigure}[b]{0.5\textwidth}   
            \centering 
            \includegraphics[width=2.8in]{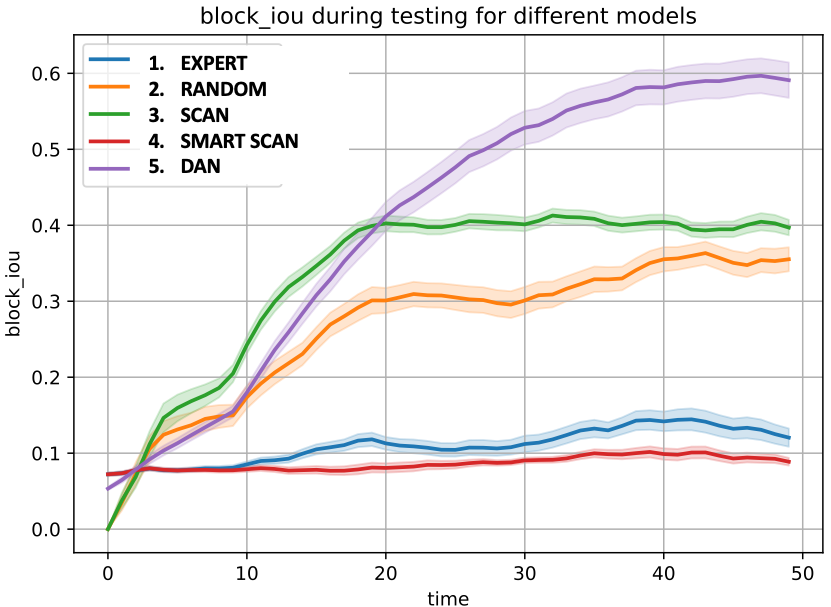}
            \caption{SpecB1 environments}
            \label{fig:man_dan_perf_specB1}
        \end{subfigure}
    \caption{\label{fig:man_dan_perf} Comparison:  hand-coded controllers expecting SpecA vs. ConvLSTM-DAN only trained in SpecA environments: (a)~testing in SpecA; (b)~testing in SpecB1. Hand-coded performance degrades significantly more than DAN when environments deviate from expected/training spec. Plot: Block IoU vs. time; 100 sampled test environments per spec.} 
\end{figure}

Fig.~\ref{fig:man_dan_perf} compares the performance of hand-coded controllers and ConvLSTM-DAN trained in only one environment spec (SpecA) and tested in multiple environment specs (SpecA and SpecB1). In 
Fig.~\ref{fig:man_dan_perf_specA} the controllers were tested in environments from SpecA. The Expert and Scan-and-Dwell controllers (which are given SpecA to estimate each sampled environment parameters) correctly predict the signal from $~t=35$. ConvLSTM-DAN achieves a good performance because it was trained on SpecA environments. Then, when tested on SpecB1 environments while being given SpecA to estimate the parameters (Fig.~\ref{fig:man_dan_perf_specB1}), the hand-coded solutions fail to correctly predict the out-of-distribution signals. In contrast, the ConvLSTM-DAN, trained only in SpecA, shows more robustness to changes in environment dynamics.

\begin{table}[tbp]
    \small
    \setlength{\tabcolsep}{2mm}
    \centering
    \begin{tabular}{l r r r}
    \toprule
    {\bf Model}         & {\bf Sep. Dense} & {\bf Shared Dense} & {\bf Shared Conv.} \\
    \midrule
    IoU                 & 0.17      & 0.25      & \textbf{0.35}      \\
    Parameters          & 82216     & 89858     & \textbf{39234}   \\
    \bottomrule
    \end{tabular}
    \caption{Our Shared ConvLSTM model converges faster to a higher accuracy (IoU) than original DAN models. Both shared models have more layers than Sep.Dense}
    \label{tab:compare-dense-conv}
\end{table}

\subsection{ConvLSTM, Predictive and InfoMax DAN}\label{sec:perf-eval:versions}

\textbf{ConvLSTM-DAN vs. original DAN.} We compare our ConvLSTM-DAN model (a shared architecture) with the models used in the original DAN: {\it Separate Dense} ($M$-net and $Q$-net are separate networks with dense layers followed by an LSTM layer) and {\it Shared Dense} (a single network with shared initial layers and separate outputs for $M$-net and $Q$-net). The ConvLSTM-DAN layers can scale arbitrarily with respect to number of frequencies without a corresponding increase in the number of model parameters and as a result it converges much faster to a better accuracy as shown in Table~\ref{tab:compare-dense-conv}.

\noindent\textbf{Predictive, InfoMax vs. ConvLSTM-DAN.} We evaluate their performance in environments with distinct challenges: (i)~Stationary (simplest): a range of patterns (period, duty-cycle) and signal numbers (1-3); (ii)~Non-stationary (more challenging and realistic): narrower range of patterns, but signal patterns and locations randomly change \textit{during} episode; (iii)~Multi-class wide spectrum environment (most realistic): multiple signal classes with different behaviours (signal modulations), aperiodic, random activity and more bands (100 vs. 20). In (iii), observation and internal state includes \textit{objectness} (is signal present) and \textit{signal class score} (softmax score vector). Also added multi-class loss, metrics and rewards.

\begin{table}[tb]
    \small
    \centering
    \begin{tabular}{l| r r r}
    \toprule
      & \multicolumn{3}{c}{\textbf{Environments}}  \\
    \textbf{Agent}  & \textbf{Stationary}     & \textbf{Non-stationary} & \textbf{Multi} \\ 
    \midrule
    ConvLSTM-DAN w/ db\_iou          & 0.62           & 0.37           & 0.50          \\
    ConvLSTM-DAN w/ in\_iou        & 0.63           & 0.38           & 0.48          \\
    Predictive-DAN        & 0.68           & \textbf{0.45}  & 0.44          \\
    Infomax-DAN         & \textbf{0.75}  & 0.39           & \textbf{0.52} \\
    \bottomrule
    \end{tabular}
    \caption{Performance of different DAN configurations and rewards, per environment (best in bold): scores are cumulative IoU, \textbf{db\_iou} is differential block IoU reward, \textbf{in\_iou} is instantaneous IoU reward (Sec.~\ref{sec:prop-soln:dan:rewards}).}
    \label{tab:enh_res}
\end{table}

Table~\ref{tab:enh_res} shows the results of our approaches.
Notice that InfoMax performs best in stationary environments, where the signal parameters vary a lot from episode to episode, but within the episode, they are stationary. Predictive-DAN is best at non-stationary environments, and RL methods (as opposed to non-RL based InfoMax) seem to perform best there. InfoMax-DAN seems to perform best at complex multi-class environments. It also shows that our designs can scale in frequency.

\subsection{Experience Feedback (Field Spec Estimation)}\label{sec:perf-eval:fieldspec}
We now cover our results in leveraging experience feedback via field spec estimation and controller retraining using the estimated spec, while field state estimation is covered in Sec.~\ref{sec:perf-eval:fieldstate}. For these experiments, our Predictive-DAN is the ML-based controller used. 

\begin{figure}[tbp]
        \centering 
           \begin{subfigure}[b]{0.5\textwidth}
        		\centering
        		\includegraphics[width=2.5in]{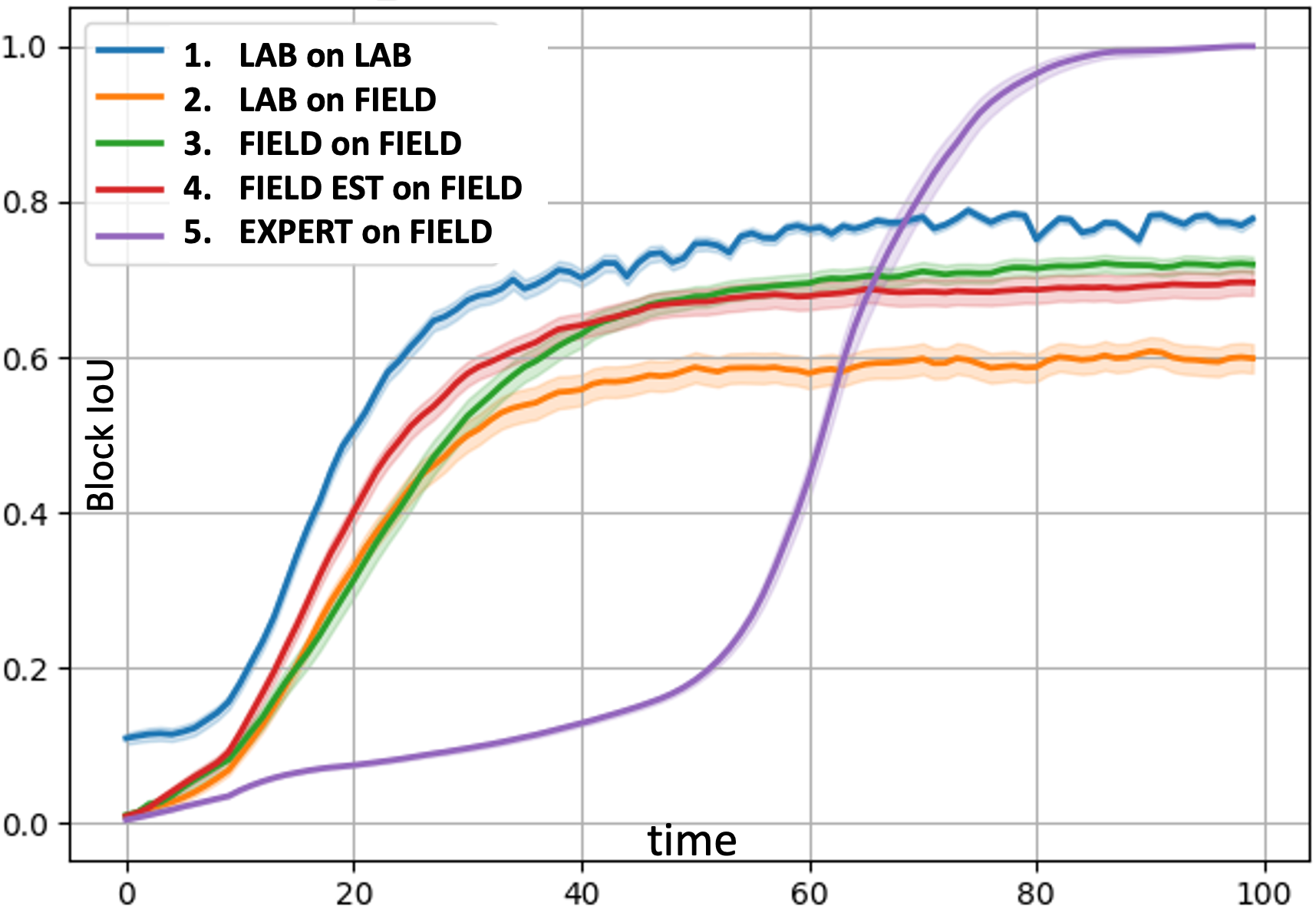}
        		\caption{}
        		\label{fig:field1_res}
        	\end{subfigure}%
        \begin{subfigure}[b]{0.5\textwidth}  
            \centering 
            \includegraphics[width=2.5in]{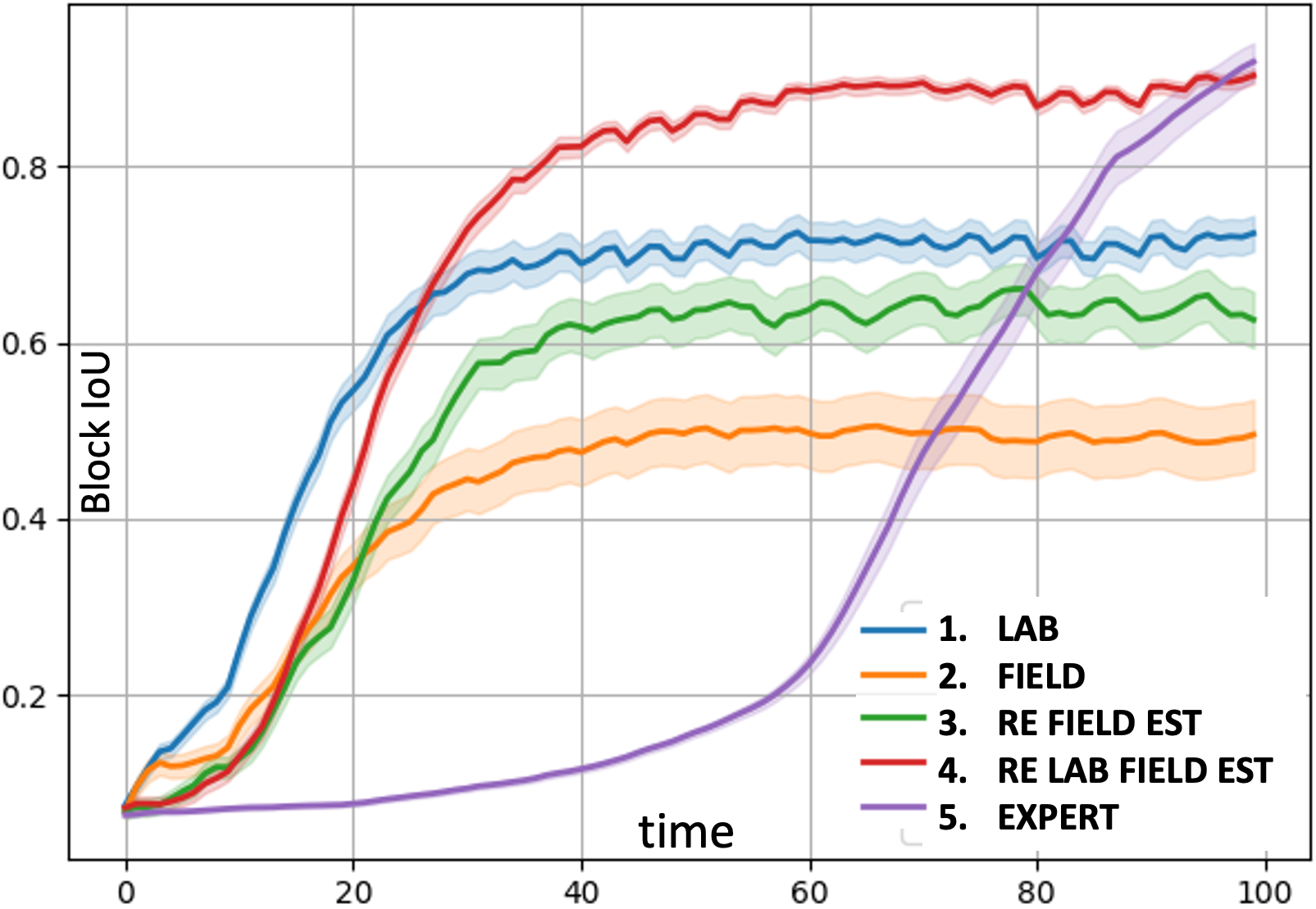}
            \caption{}
            \label{fig:field2_res}
        \end{subfigure}
        \caption{\label{fig:field1_2_res} Test results plotting timesteps vs. Block IoU 
        comparing: (a)~Estimating Spec from Feedback. 
        (b)~Estimating Spec from Retraining. (Block IoU of 5 steps.) 
        }
\end{figure}

\noindent\textbf{Estimating Spec from Feedback.}
In Fig.~\ref{fig:field1_res} we compare our approach (4.~\textit{Field Est on Field}) to a scan-and-dwell hand-coded control where field spec is within-distribution (5.\textit{Expert on Field}) and also to our ML controller alternatively trained (1.-3.). The hand-coded controller takes a long time to reduce uncertainty (large search space). A DAN controller trained on lab alone cannot cope with field (2.~\textit{Lab on Field}). By learning estimated field spec \emph{without} GT, our approach has similar performance to training on field spec \emph{with} GT (3.\textit{Field on Field}, see also analogous 1.\textit{Lab on Lab}).

\noindent\textbf{Estimating Spec from Retraining.}
In Fig.~\ref{fig:field2_res}, we compare a controller (4.~\textit{Re Lab Field Est}) trained on Lab spec, then re-trained on a mix of estimated field spec ($0.7$ weight) + Lab spec ($0.3$ weight) to a scan-and-dwell hand-coded controller (5.~\textit{Expert}) and also to our ML controller alternatively trained (1.-3.). For this experiment, Field Spec and Lab Spec differ in number of signals (1 vs. 2, respectively) and width between signals (3 vs. 2), and we sample a mix of both specs for testing. Controllers trained on only one spec (1.~\textit{Lab}, 2.~\textit{Field}) cannot cope with environments from multiple specs. The controller retrained only on estimated field spec (3.~\textit{Re Field Est}) suffers from ``forgetting'' and cannot cope with environments from it's initial training spec. By retraining on both specs (weighted), the controller has the best performance.

\begin{figure}[tbp]
        \centering 
        \hspace{-0.3in}
~
~
            \includegraphics[width=2.5in]{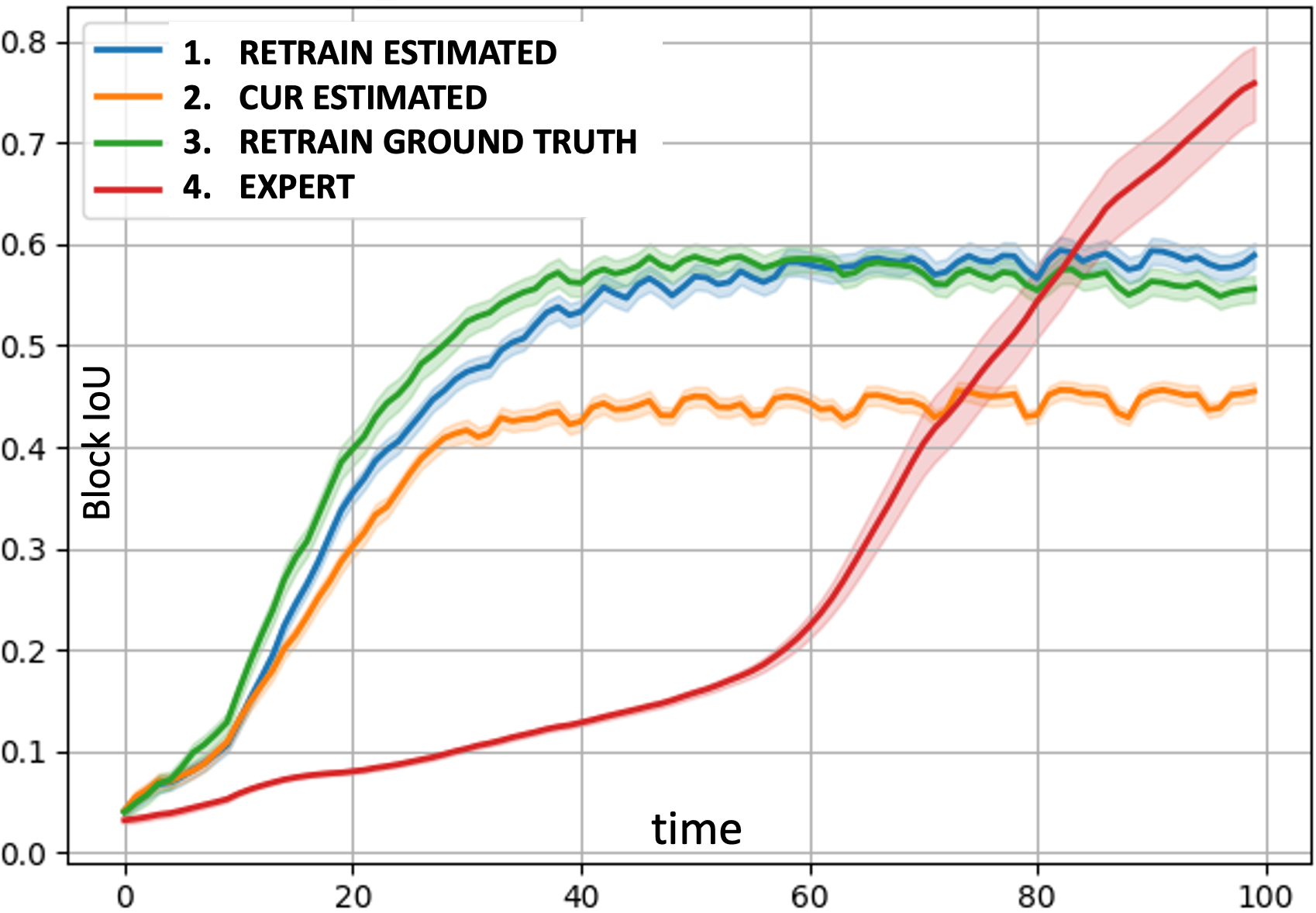}
        \caption{\label{fig:field3_res} Comparison of DAN vs Expert controllers for the bootstrapping approach: performance on 3\textsuperscript{rd} (last) iteration shown. (Block IoU of 5 steps.)
        }
\end{figure}

\noindent\textbf{Boostrapping.}
In Fig.~\ref{fig:field3_res}, we consider a bootstrap loop with 3 iterations, where the ground-truth field spec parameters differ slightly from the previous iteration's (lab spec used is A and the three field specs are F1-F3 in Table\ref{tab:envs}). The idea is to expose a controller to environments of variable complexity. The results for the last iteration are shown in Fig.~\ref{fig:field3_res}, where we evaluate against \textit{all} environment specs (lab + 3 field). The detection rate ($\approx 60\%$) of the controller with estimated specs (1.~\textit{Retrain Estimated}) is close to that of controller using GT specs (3.~\textit{Retrain Ground Truth}) $\approx 65\%$. 
Hence, training with GT loses advantage over training with estimated specs over multiple iterations as the agent is exposed to more and more types of environments. We can also see that re-training DAN controller leads to smooth adaptation and using only current estimated spec (2.\textit{Cur Estimated}) results in poor performance.

\new{
\subsection{Experience Feedback (Field State Estimation)}\label{sec:perf-eval:fieldstate}

Instead of estimating the parameters of a field spec, which is then used to train a controller, we can perform \emph{Field State Estimation}: do ML-based estimation of the full field/spectrum state (2. in Fig.~\ref{fig:model_based1}) from partially observed field observations (1. in Fig.~\ref{fig:model_based1}), and then use these reconstructed full-state episodes directly to retrain our ML controller, as described in Sec.~\ref{sec:prop-soln:exp-fb:state}. The ML-based estimation is done by a generator that is trained offline, separately from DAN. Once trained, it then generates one full-state episode from one partially observed field episode.
In Table.~\ref{tab:model_based_res}, we compare the field state estimation approach to a controller exhaustively trained on an (anticipated) lab spec and one trained on the ideal training source: the field spec with complete ground truth information. We also study the performance after training with different numbers of reconstructed full-state episodes (25, 50, 100). Finally, we compare the approach with fine-tuning a controller in the field alone using partially observed prediction rewards---reward only what you observe in the band sampled and no error information using GT in other bands. For these experiments, we use \textit{SpecC1} (Table~\ref{tab:envs}) for the lab spec and \textit{SpecC2} for the field spec. Throughout these experiments, evaluations of all training configurations are done on the same 100 field spec episodes (not used in training) and Predictive-DAN is the controller used.

\begin{table}[tb]
    \small
    \centering
    \begin{tabular}{l| r}
    \toprule
    \textbf{Training approach}  & \textbf{Block IoU} \\ 
    \midrule
    1 Trained with lab spec only        & 0.58 \\
    2 Retrained w/ 25 estimated field state episodes  & 0.53 \\
    3 Retrained w/ 50 estimated field state episodes & 0.70 \\
    4 Retrained w/ 100 estimated field state episodes & 0.72 \\
    5 Training w/ field spec and full GT  & 0.75 \\
    6 Finetuned w/ 100 partially observed field episodes & 0.43 \\
    \bottomrule
    \end{tabular}
    \caption{Experience feedback using \emph{Field State Estimation} (rows 2-4), compared with training on (1) lab only, (5) field with full GT and (6) partially observed field episodes. Block IoU over last 33 steps.}
    \label{tab:model_based_res}
\end{table}

The first row in Table.~\ref{tab:model_based_res} corresponds to DAN trained with the lab spec, with no field experience and 1000 unique episodes. We can see that the performance is poor. The next three rows (2-4) correspond to DAN retrained with $25/50/100$ estimated full-state field episodes mixed with $25/50/100$ lab spec episodes, respectively.  We can see that performance improves when the lab controller is retrained with more estimated field episodes. The fifth row in Table.~\ref{tab:model_based_res} corresponds to training from the field spec with full ground truth and for 1000 unique episodes. This result roughly corresponds to the best possible performance of DAN given GT. As we can see, in this case, retraining DAN with 100 \emph{estimated} field episodes (using no GT) 
has similar performance to training DAN with field GT and 1000 episodes.

The result of evaluating the DAN controller fine-tuned directly on 100 \textit{partially observed field episodes} alone is shown in the last row (6) of Table.~\ref{tab:model_based_res}. For this case, we modified training for DAN accordingly to accommodate partially observed states, i.e., we have ground truth information only in the particular band sampled instead of all the bands previously---hence the reward obtained by $M$-network is limited to the prediction only in this band. We can see that its performance is inferior compared to training with full-state episodes, even when those complete states are estimated (row 4 of Table.~\ref{tab:model_based_res}). Partial observations in sparse, dynamic signal environments makes training difficult.

In summary, we show that retraining a DAN with full-state sequences estimated using ML-based generators from partially observed field observations is a viable and promising approach, while fine-tuning a DAN \emph{directly} on partially observed field experiences produces 
poor results in general.
}

\section{Conclusion and Future Work}\label{sec:conc}

Novel RL-based approaches for information gain under partially observable, non-stationary and reward-sparse environments are presented and applied to the complex problem of RF spectrum monitoring, where the task is to identify signals with distinct behavioral patterns at variable frequencies. The proposed solution accounts for scalability and translation-invariance challenges of the signals in frequency-time space. We also propose information gain rewards to encourage exploration of unseen signals. 
Additionally, we developed two model-based RL approaches for the difficult task of retraining/fine-tuning a lab-trained controller using experience feedback from limited field deployment without ground-truth. 
Simulation results indicate that our approach outperforms previous RL and hand-coded solutions, and is promising for monitoring RF spectrums. We also show promising results for retraining our ML controller using limited field experience.

For future work, we plan on studying adversarial environments with competitive multi-agent learning and complex control surfaces with action representation learning.
\new{Also, we plan to investigate the use of GANs to generate training sequences from sparse field samples.}

\section*{Acknowledgements}
This material is based upon work supported by the Defense Advanced Research Projects Agency (DARPA) and Space and Naval Warfare Systems Center, Pacific (SSC Pacific) under Contract No. N66001-18-C-4044. 

\textbf{Disclaimer}: The views, opinions, and/or findings expressed are those of the author(s) and should not be interpreted as representing the official views or policies of the Department of Defense or the U.S. Government.


\bibliographystyle{unsrtnat}
\bibliography{refs}

\end{document}